\pdfoutput=1
\documentclass[11pt]{article}

\usepackage[]{EMNLP2023}

\usepackage{times}
\usepackage{ragged2e}
\usepackage{latexsym}
\usepackage{caption}
\usepackage{subcaption}
\usepackage{graphicx}
\usepackage{subfloat}
\usepackage{booktabs} % To thicken table lines
\usepackage{multirow}
\usepackage{amsmath,amssymb}
\usepackage{cuted}
\usepackage{flushend}
\usepackage{tablefootnote}
\usepackage{caption}
\usepackage{subcaption}
\usepackage{graphicx}
\usepackage{subfloat}
\usepackage{wrapfig}
\usepackage[T1]{fontenc}
\usepackage[utf8]{inputenc}
\usepackage{pifont}
\usepackage{microtype}
\usepackage{inconsolata}
\usepackage{xcolor}
\usepackage{booktabs}

\usepackage[framemethod=TikZ]{mdframed}
\usepackage{array} 
\usepackage{tabularx, booktabs}
\usepackage{longtable} % Include the longtable package
\usepackage{hyperref}

\newmdenv[%
    backgroundcolor=gray!10,
    linecolor=black,
    outerlinewidth=0.5pt,
    roundcorner=1mm,
    skipabove=\topsep,
    skipbelow=\topsep,
    font=\ttfamily\tiny,
]{promptbox}

\usepackage{todonotes}
\usepackage{xcolor}

\usepackage{makecell}

\title{Beyond Metrics: Evaluating LLMs' Effectiveness in Culturally Nuanced, Low-Resource Real-World Scenarios}

\author{\parbox{0.9\linewidth}{\centering{Millicent Ochieng\textsuperscript{$\dagger$} \quad Varun Gumma\textsuperscript{$\dagger$} \quad Sunayana Sitaram\textsuperscript{$\dagger$} \quad Jindong Wang\textsuperscript{$\dagger$} \quad Vishrav Chaudhary \textsuperscript{$\dagger$} \quad Keshet Ronen\textsuperscript{$\diamondsuit$} \quad Kalika Bali\textsuperscript{$\dagger$} \quad Jacki O'Neill\textsuperscript{$\dagger$} \\
 {\rm \textsuperscript{$\dagger$}Microsoft Corporation \textsuperscript{$\diamondsuit$}University of Washington \\}
 {\tt Contact: {\{mochieng, jacki.oneill\}@microsoft.com}}}}}

\begin{document}

\maketitle
\begin{abstract}
The deployment of Large Language Models (LLMs) in real-world applications presents both opportunities and challenges, particularly in multilingual and code-mixed communication settings. This research evaluates the performance of seven leading LLMs in sentiment analysis on a dataset derived from multilingual and code-mixed WhatsApp chats, including Swahili, English and Sheng. Our evaluation includes both quantitative analysis using metrics like F1 score and qualitative assessment of LLMs' explanations for their predictions. We find that, while Mistral-7b and Mixtral-8x7b achieved high F1 scores, they and other LLMs such as GPT-3.5-Turbo, Llama-2-70b, and Gemma-7b struggled with understanding linguistic and contextual nuances, as well as lack of transparency in their decision-making process as observed from their explanations. In contrast, GPT-4 and GPT-4-Turbo excelled in grasping diverse linguistic inputs and managing various contextual information, demonstrating high consistency with human alignment and transparency in their decision-making process. The LLMs however, encountered difficulties in incorporating cultural nuance especially in non-English settings with GPT-4s doing so inconsistently. The findings emphasize the necessity of continuous improvement of LLMs to effectively tackle the challenges of culturally nuanced, low-resource real-world settings and the need for developing evaluation benchmarks for capturing these issues. 
\end{abstract}
\section{Introduction}

Large Language Models (LLMs) have ushered in major advancements in language processing, demonstrating exceptional ability to process everyday language commands and handle textual tasks such as Question Answering, Sentiment Analysis, Summarization, among others \cite{OpenAI2023,Brown2020,Chowdhery2022,geminiteam2023gemini, Touvron2023}.

Despite LLMs advancements, their effectiveness is predominantly observed in Latin Script languages with abundant training data, such as English, which constitutes a significant proportion of their training corpus \cite{JMLR:v21:20-074,CommonCrawlLangStats2023,together2023redpajama,longpre2023data}. Although English is not the mother tongue of the majority of the world's population, 93\% of GPT-3's training data consists of  English content  \cite{Brown2020}. Studies reveal that languages with medium to low amounts of training data like Swahili still present challenges for these models, highlighting they are far from achieving parity with English  \cite{ahuja-etal-2023-mega,  Ahuja2023MEGAVERSEBL, robinson-etal-2023-chatgpt}. The picture is further complicated given that 60\% of the world population speaks two or more languages\footnote{\url{https://ilanguages.org/bilingual.php}}. In such settings, code-mixing\footnote{the practice of alternating between two or more languages or dialects in a conversational turn} is a prevalent aspect of natural language use.  Consequently, the performance of these models in real-world settings, especially in low-resource code-mixed and culturally diverse environments, remains an area of significant interest. 

This study investigates the effectiveness of seven prominent LLMs on a sentiment analysis task on a dataset derived from WhatsApp chats. The dataset exhibits extensive code-mixing, encompassing multilingual conversations in English, Swahili, and Sheng\footnote{a dynamic urban slang from Nairobi, Kenya, blending Swahili, English, and local languages, evolving continually among the youth.} in `chat speak' e.g. using emojis, abbreviations, colloquial chat message spellings and mispellings. With LLMs' ability to process and produce human-like text, this task aims to evaluate their understanding of the nuances present in the dataset. We supplement the quantitative analysis of the LLMs performance with a systematic qualitative analysis of the explanations the models provide for their predictions. While studies such as \cite{Narang2020WT5TT, wiegreffe-etal-2021-measuring, Majumder2021KnowledgeGroundedSV, wiegreffe-etal-2022-reframing} have demonstrated the capability of LLMs to generate natural language explanations alongside predictions, enhancing explainability and improving the faithfulness of AI systems, it remains uncertain whether these explanations directly influence the decision-making process. However, we expect, and indeed do see, a correlation between models' predictions and their explanations. We used the explanations as a method of interrogating, to some extent, the models ability to process the cultural and linguistic nuances of the messages. By looking beyond the numbers, this method enables us to get some sense of how well the different LLMs handle the complex interactional features present in a real-world multilingual dataset. We demonstrate the value of using qualitative HCI methods alongside traditional performance metrics. 
Our contributions are as follows: (1) we evaluate and compare the performance of seven advanced LLMs including GPT-4, GPT-4-Turbo, GPT-3.5-Turbo, Llama-2-70b, Mistral-7b, Mixtral-8x7b and Gemma-7b on a sentiment analysis task using a novel WhatsApp chat dataset; (2) we identify differences in the interpretation strategies employed by different LLMs, highlighting the diversity in their approach to processing complex linguistic data; (3) we highlight the value of real-world, multilingual, and code-mixed datasets in assessing the performance of LLMs; (4) we show how qualitative HCI methods can be used in NLP to get a deeper understanding of model performance. 
Our findings reveal that, while LLMs like Mistral-7b and Mixtral-8x7b achieved high F1 scores in sentiment analysis in the dataset, they and other LLMs such as GPT-3.5-Turbo, Llama-2-70b, and Gemma-7b seem to be less robust at handling linguistic, cultural, and contextual nuances. Further, there was a lack of transparency in their generated explanations. In contrast, LLMs like GPT-4 and GPT-4-Turbo deployed diverse linguistic and contextual information in their explanations, demonstrating high consistency with human judgement. All the LLMs however, struggled to incorporate the more complex cultural nuances in the WhatsApp dataset especially in non-English settings - even GPT-4 and GPT-4-Turbo did so inconsistently.
% \newpage
\section{Evaluation Dataset and Task}

\subsection{Dataset}

\paragraph{The WhatsApp Chat Dataset:} Our study employed a distinctive dataset originally collected by \citet{Karusala2021ThatCT} further annotated by \citet{mondal-etal-2021-linguistic}, with all ethical considerations and privacy measures observed as described below.  It features multilingual exchanges among young people living with HIV in informal settlements in Nairobi, Kenya, captured within two health-focused WhatsApp chat groups moderated by a medical facilitator. The total number of messages are 6,556 and the conversations are predominantly in English, enriched with a considerable use of Swahili, Sheng, and code-mixing. The data annotation included sentiment and word-level language identification for each message. As \citet{Karusala2021ThatCT} describe, recruited participants signed a consent form outlining study procedures, data anonymization, and security measures. All messages were anonymized and translated into English by a native speaker. Each chat message in the dataset included an anonymized speaker ID, timestamp, original message, and English translation. Due to the sensitive nature of the content, the dataset is not publicly available, but researchers can request access by contacting the authors. We specifically selected this dataset because it consists of real WhatsApp interactions between participants and a medical facilitator occurring as part of a Global Health research intervention. Additionally, its authentic representation of real-world, code-mixed communication aligns with our core research focus.

\paragraph{Pre-Processing:} Considering that the data originates from WhatsApp conversations, it exhibits a casual, conversational style, often with short interactions. We retained only turns with three or more words providing more valuable data for sentiment analysis. Contrary to typical processing methods, we do not perform punctuation or emoji normalization on the data, as these elements are integral to the communication. The resulting dataset consisted of 3,719 messages with an average of eleven words per message.

 \begin{table*}[tp]
     \centering
     \tiny
     \begin{tabular}{|m{1.9cm}|m{1.3cm}|m{1.4cm}|m{1.4cm}|m{1.8cm}|m{2.5cm}|m{2cm}|}
     \toprule
    \textbf{Language} &  \textbf{\# of Neutral Messages}&  \textbf{\# of Positive Messages} &  \textbf{\# of Negative Messages} &  \textbf{\# of Messages Per Language} & \textbf{Average \# of Tokens Per Message} & \textbf{Total \# of Tokens Per Message}\\
    \midrule
    \textbf{\textit{Monolingual}} \\
    \midrule
    En	& 1303 & 54 & 24 & 1381 & 12 & 16902\\
    Sw	& 270 & - & 32 & 302 & 4 & 1264\\
    Sh	& 2 & - & - & 2 & 3 & 6\\
    \midrule
    \textbf{\textit{Multilingual}} \\
    \midrule
    En-Sw & 631 & 5 & 19 & 655 & 9 & 5582\\
    Sw-Sh & 143 & - & 12 & 155 & 5 & 705\\
    En-Sh & 51 & 2 & 1 & 54 & 6 & 301\\
    En-Sw-Sh & 190 & 3 & 16 & 209 & 10 & 2100\\
    En-CM & 10 & - & 1 & 11 & 9 & 94\\
    Sw-CM & 29 & - & 2 & 31 & 4 & 124\\
    En-Sw-CM & 60 & - & 6 & 66 & 12 & 812\\
    En-Sh-CM & 2 & - & - & 2 & 12 & 24\\
    Sw-Sh-CM & 20 & 1 & 1 & 22 & 5 & 106\\
    En-Sw-Sh-CM & 36 & - & 3 & 39 & 14 & 542\\
    Other & 96 & 1 & - & 98 & 20 & 795\\
    En-Other & 359 & 10 & 4 & 373 & 20 & 7622\\
    Sw-Other & 73 & - & 5 & 78 & 4 & 342\\
    Sh-Other & 1 &  & - & 1 & 4 & 4\\
    En-Sw-Other & 119 & 2 & 2 & 123 & 11 & 1396\\
    En-Sh-Other & 16 & - & 2 & 18 & 10 & 188\\
    Sw-Sh-Other & 19 & - & 1 & 20 & 6 & 110\\
    En-Other-CM & 4 & - & - & 4 & 6 & 24\\
    Sw-Other-CM & 4 & - & 3 & 7 & 6 & 41\\
    En-Sw-Sh-Other & 37 & 2 & 1 & 40 & 12 & 481\\
    En-Sw-Other-CM & 10 & - & 2 & 12 & 13 & 161\\
    En-Sh-Other-CM & 3 & - & - & 3 & 4 & 12\\
    Sw-Sh-Other-CM & 4 & - & - & 4 & 8 & 33\\
    En-Sw-Sh-Other-CM & 9 & - & - & 9 & 17 & 151\\
    \midrule
    Total & \textbf{3501} & \textbf{80} & \textbf{137} & \textbf{3719} & \textbf{246} & \textbf{39922}\\  
  \bottomrule
     \end{tabular}
     \caption{Message Distribution by Language. This table displays the count of neutral, positive, and negative messages, total messages per language, average tokens per message, and total tokens per message for each language studied.}
     \label{tab:BroadLanguageCount}
 \end{table*}

\subsection{Evaluation Task}

\paragraph{Sentiment Analysis:} The core of our evaluation focuses on a sentiment detection task because of its real-world application for such chat groups. We wished to support the facilitator by for example flagging negative messages. Table \ref{tab:BroadLanguageCount} illustrates the sentiment distribution according to human annotators within our pre-processed dataset, heavily skewed towards the Neutral class. This imbalance highlights the evaluation challenge of accurately identifying the less frequent Negative and Positive sentiments, testing the LLMs' ability to detect sentiment cues in a predominantly Neutral context.

\paragraph{Languages in the WhatsApp Dataset:} Table \ref{tab:BroadLanguageCount} describes the statistics of languages within the dataset defined as: \textit{En (English), Sw (Swahili), Sh (Sheng) and CM (Code-Mixed)}. These include messages in single language (\textit{Monolingual}) and messages in more than one language (\textit{Multilingual}). The dataset includes an `Other' category used for words that do not fit the primary categories due to uncertainty, named entities, or other unique factors.
\section{ Experimental Setup}
 
\subsection{Models}

We evaluated three OpenAI models: \texttt{GPT-4-Turbo}, \texttt{GPT-4-32k} \cite{gpt4techreport}, and \texttt{GPT-3.5-Turbo} \cite{Ouyang2022TrainingLM}, with \texttt{GPT-4-32k} being the latest iteration and known for its enhanced performance on text processing and generation. \texttt{GPT-4-Turbo} and \texttt{GPT-3.5-Turbo} are optimized versions designed for more efficient processing without significantly compromising performance. From the open-source collection, we select Meta's \texttt{Llama-2-70b-chat} \cite{Touvron2023}, an LLM known for its efficiency and chat functionality. Additionally from Mistral AI, we include \texttt{Mistral-7B-Instruct-v0.2} \cite{jiang2023mistral} and \texttt{Mixtral-8x7B-Instruct-v0.1} \cite{Jiang2024MixtralOE}, the former being popular for its exceptional ability to follow instructions and the latter for its innovative architecture which makes it excel in mathematics, code generation, and multilingual tasks. Lastly, we include Google's \texttt{Gemma-7b-it} \cite{Mesnard2024GemmaOM}, a state-of-the-art language model that excels in language understanding, reasoning, and safety, outperforming comparable models in numerous academic benchmarks. Throughout this paper, we refer to the mentioned models as: GPT-4, GPT-4-Turbo, GPT-3.5-Turbo, Llama-2-70b, Mistral-7b, Mixtral-8x7b and Gemma-7b.

\subsection{Model Evaluation}

Different prompting approaches \cite{Brown2020,chen2023unleashing} have been shown to effectively guide LLMs contextually, towards desired outputs. Investigations reveal that the quality of prompts provided, have a profound influence on the performance of LLMs \cite{10.1145/3560815,Hada2023large}. Leveraging this technique, we craft a detailed prompt to guide the LLMs to function as specialized NLP assistants for sentiment analysis. The prompt directs models to identify sentiments as Positive, Negative, or Neutral. Figure \ref{fig:prompt_here} illustrates the standardized prompt that we used for evaluating all seven LLMs, facilitating a fair and consistent comparison of their performance. 
We employ the same sentiment definitions given to human annotators during the dataset's sentiment annotation phase. Furthermore, we direct the models to justify their sentiment classifications in 200 words or less, focusing on the text spans that influenced their decisions. We conduct the evaluation on the entire pre-processed dataset and employ the weighted F1-score\footnote{\url{https://scikit-learn.org/stable/modules/generated/sklearn.metrics.f1_score.html}} metric instead of accuracy due to the skew in our dataset.

\section{Qualitative Analysis}

We supplemented our quantitative evaluation with in-depth qualitative analyses. Considering the skew towards the neutral class in our dataset, as illustrated in Table \ref{tab:BroadLanguageCount}, and to ensure a balanced and rigorous analysis, we selected a sample comprising a total of 261 messages at random including both monolingual and multilingual messages for in-depth examination. For monolingual, the sample included 150 messages divided equally among the three sentiment categories: Negative, Positive, and Neutral. For the multilingual case, the sample entailed 50 Neutral, 50 Negative and 11 Positive messages whereby the positive messages represented all the positive labels in this category. The first author being proficient with the languages in the WhatsApp dataset, analyzed all the 261 sampled messages, identifying patterns in the data. We borrowed five criteria as outlined by \cite{Chang2023ASO} on \textit{`how to evaluate'} to guide our human evaluation. These criteria, described in Table \ref{tab:CriteriaDescription}, were linguistic accuracy, contextual and cultural relevance, fluency in maintaining consistency, alignment with human expectations and transparency in the LLMs' decision-making process. While these criteria have been designed for quantitative evaluation, we used them to provide a structure for analysing the set of justifications for each message across models. We supplemented this structured analysis with a more in-depth ethnomethodologically-informed approach \cite{6dbb16182798488ebe2c1bf86796b326, Crabtree2006ThePI, 10.1145/958160.958167, c6b7bbf5dc6d41308ed6147323139c14} where the first author and the last author (who is skilled in ethnomethodologically-informed analysis) together analysed each turn (message) in detail, to understand how the justifications produced by the different models related to the original and the human-translated message, how they related to one another and how they related to the sentiment prediction. In these sessions, the two authors looked in detail at the messages and model justifications and identified emergent patterns, interrogating and refining them. This analysis was deeply qualitative,  aiming to derive insights into differences between models in their justifications. As an additional sanity check, we invited three other native speakers to review a set of 15 messages, selected from the 261 messages the first author had analysed. The reviewers conducted their assessments independently and reconvened to discuss their findings, along with the first author and were all confident about the consistency of the findings. In this paper, for reasons of space, we use a small number of examples to illustrate the patterns that we found in the data. 
\section{Results and Discussion}

In this section, we explore the results of our study by discussing both quantitative and qualitative findings of models performance on the WhatsApp dataset, beginning with the quantitative results measured by the F1 score. Following this, we will delve into the qualitative findings, discussing insights from the models' justifications and their implications for language processing strategies.

\subsection{F1 Score-Based Models Comparison}

\begin{figure}[h!]
    \centering
    \includegraphics[width=\linewidth]{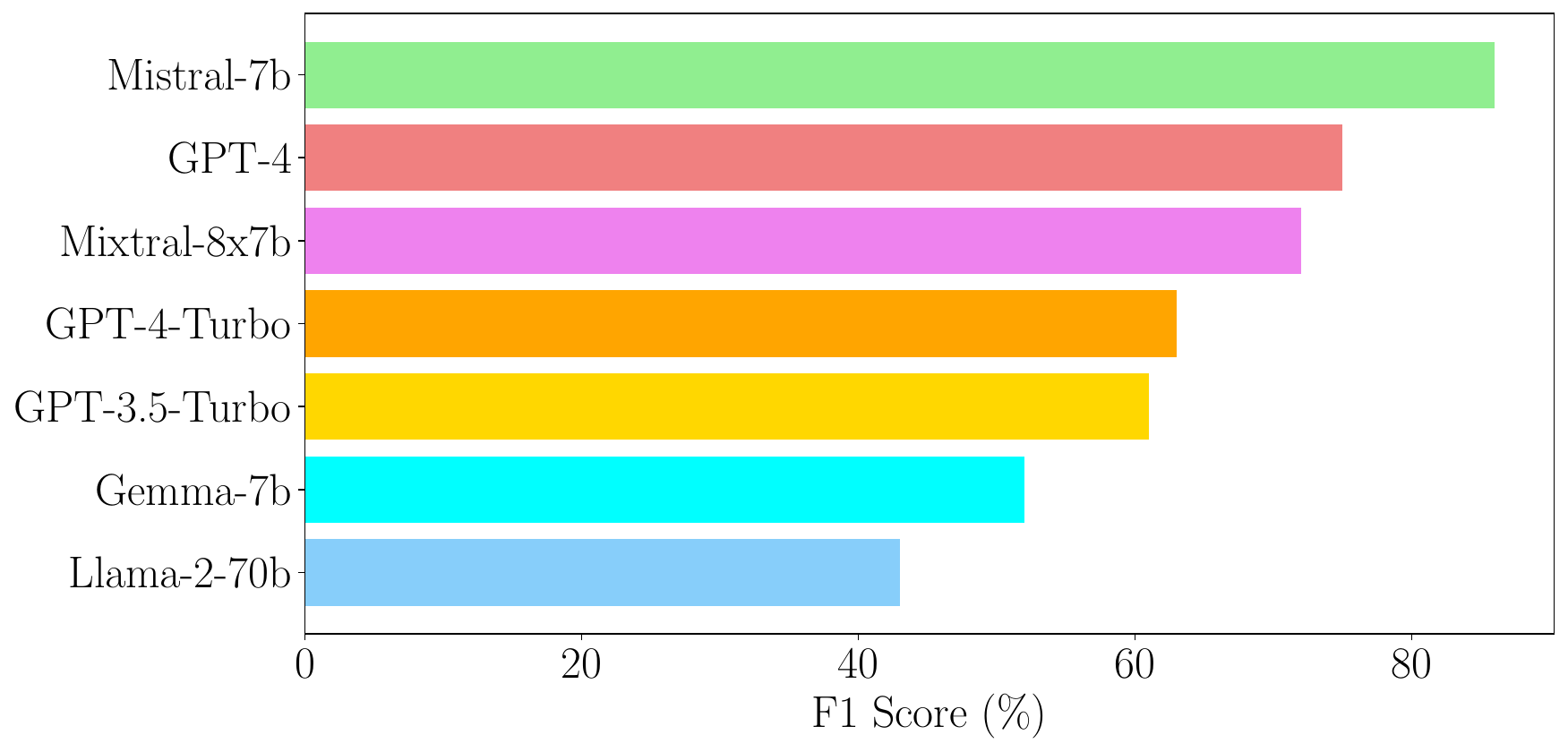}
    \caption{Overall F1 score comparison of the models.}
    \label{fig:overall_f1_scores}
\end{figure}

As evidenced by the F1 Score comparison of the seven models in Figure \ref{fig:overall_f1_scores}, the Mistral-7b model demonstrates a higher performance in sentiment analysis on the WhatsApp dataset, closely followed by GPT-4. Conversely, the Llama-2-70b model exhibits the weakest performance.

\begin{figure}[h!]
    \centering
    \includegraphics[width=\linewidth]{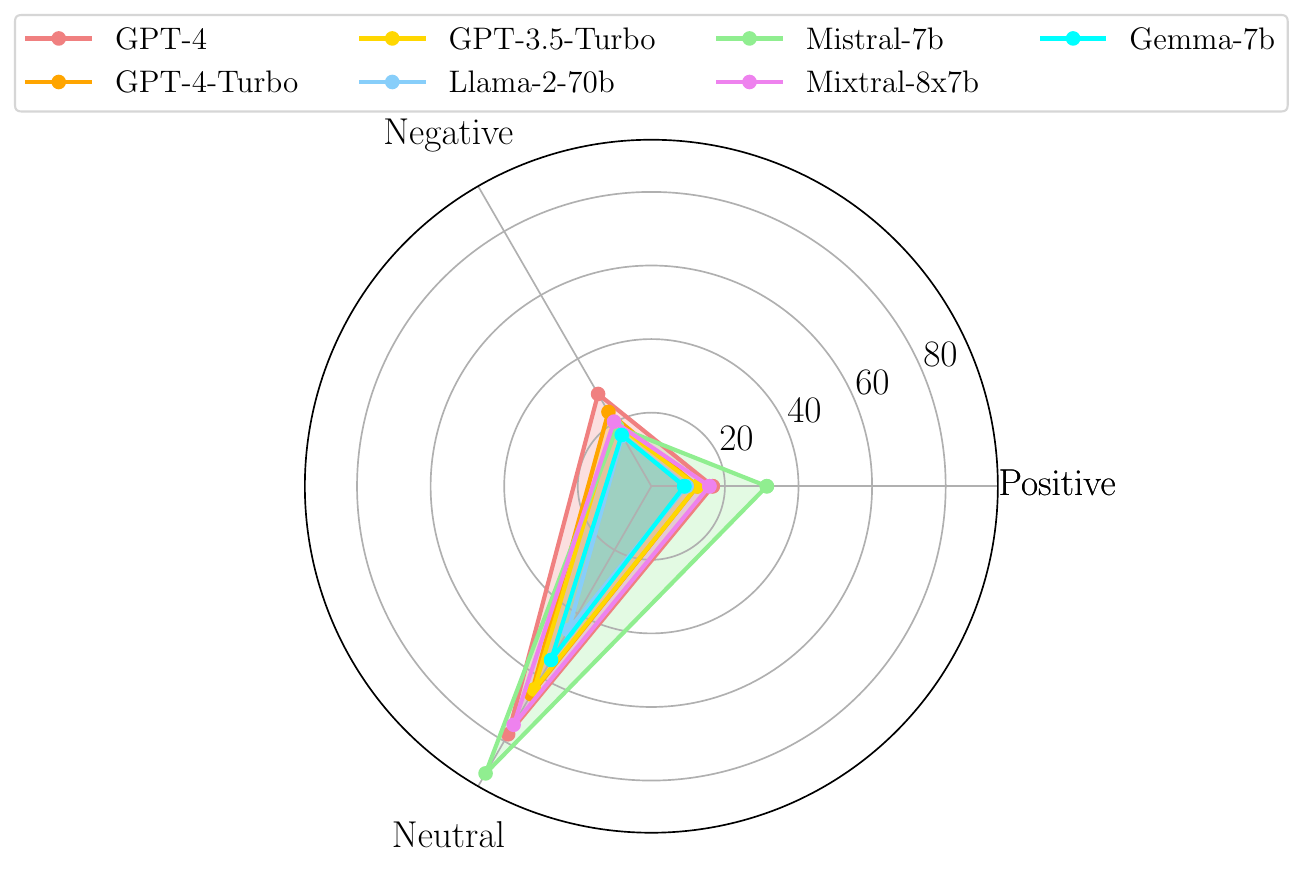}
    \caption{Comparison of F1 scores for the models across Positive, Negative, and Neutral sentiments.}
    \label{fig:different_class_overall_f1_scores}
\end{figure}

As Table \ref{tab:BroadLanguageCount} illustrates, the majority of positive and neutral sentiments were expressed in English, whereas most negative sentiments were conveyed in Swahili. This supports the findings of \cite{rudra-etal-2016-understanding}, which suggest that people are more likely to express negative opinions in non-English languages.
Further analysis of F1 scores by sentiment and language, as shown in Figures \ref{fig:different_class_overall_f1_scores} and \ref{fig:f1_scores_by_languages}, highlights the distinct capabilities of various models. Specifically, Mistral-7b excels in identifying neutral sentiments, predominantly in English followed closely by GPT-4s, Mixtral-8x7b, and GPT-3.5-Turbo, with Gemma-7b and Llama-2-70b trailing. Conversely, GPT-4 and GPT-4-Turbo demonstrate superior performance in accurately classifying the rare negative sentiments, predominantly in Swahili and code-mixed. These findings are consistent with those from standard NLP benchmarks such as those reported in \cite{Ahuja2023MEGAVERSEBL}, particularly in non-English contexts, specifically low-resource languages like Swahili and code-mixed languages. In these settings, larger models such as OpenAI's GPT-4s frequently outperform other LLMs.

% Sentiment

\subsection{Insights from the Justifications}

As instructed by our detailed prompt in Figure \ref{fig:prompt_here}, all the seven LLMs produced their predictions along with justifications. Across the board, even where sentiment predictions were the same for all the models, we noticed distinct differences in the justifications provided. Some LLMs consistently incorporated words or spans of text as part of their justifications (as requested in the prompt) others did so less frequently. Similarly some regularly translated non-English terms into English, others did not. See the examples presented in this Section's Tables \ref{tab:example_messages1} and \ref{tab:example_messages2}
 and in Appendix \S\ref{sec:appendix_analysis}.  In Table \ref{tab:example_messages1}, GPT-3.5-Turbo and Mistral-7b do not provide any spans of message text in their justifications.

\textbf{Which models perform best at accurately interpreting linguistic nuances and textual inaccuracies such as spelling errors, local abbreviations and grammatical inaccuracies?} As illustrated by the examples in Tables \ref{tab:example_messages2} and \ref{tab:example_messages4} and Appendix \S\ref{sec:appendix_analysis}, GPT-4 and GPT-4-Turbo stand out with superior performance in languages like Swahili and Sheng, and in code-mixed scenarios where they effectively handle the linguistic nuances between English and local languages. In particular, they were more consistently correct in their \textbf{translation} of the non-English word spans used in their justifications. They maintained strong performance even in rare sentiment classes, outperforming models like GPT-3.5-Turbo, which, though proficient, does not reach the high level of linguistic performance exhibited by the GPT-4s. Taking the example in Table \ref{tab:example_messages3} to illustrate - a code-mixed English-Swahili-Sheng message, GPT-4 and GPT4-Turbo provide correct interpretations of the Swahili and Sheng in their justifications, with GPT-4 Turbo even identifying \textit{`kuniboo'} (Translation: \textit{`bore me'}) as Sheng. Gemma-7b also identifies Sheng, but wrongly identifies the whole sentence as Sheng, and mistranslates it. The other models all provide mistranslations in their justifications. Models such as Llama-2-7b, Mistral-7b, Mixtral-8x7b and Gemma-7b face difficulties with Swahili and Sheng, as evidenced by their often incorrect translations of Swahili words and phrases. They often \textbf{prioritized English} in mixed-language settings resulting in either incorrect predictions or justifications when key sentiment indicators lie in the non-English segments which was mostly the case for Negative sentiments. Similarly with regards to LLMs' robustness to textual inaccuracies, the GPT-4s accurately interpreted messages with irregularities. However, the remaining models were less reliable, struggling with noisy data especially non-English texts; an example is shown in Table \ref{tab:example_messages4}.

Nonetheless, overall all the seven LLMs have demonstrated proficiency in English messages in the WhatsApp dataset. However, even in English the models can fail to predict the correct sentiment, and their justifications  reflect the sentiment that they predicted. Let's take the example in Appendix Table \ref{tab:example_messages5}, for a human reviewer, this message is a clear example of a social media chain message - typically 'copy and paste' messages requiring the reader to either like, respond, or forward. The linguistic indicators of this are the instruction at the start \textit{"Send to everyone you love..."} and the conclusion \textit{"You are lovable if you get FIVE sent back to you"}. This can be read as an instruction or perhaps a playful activity.  None of the models predicted the correct sentiment (Neutral), all predicting Positive. In their justifications, none of them identified this as a chain message, instruction, activity or  similar - even where they highlighted the phrase \textit{"You are lovable if you get FIVE"}. 

\textbf{Do the models utilize the surrounding textual context and cultural subtleties to determine sentiment?}
From our analysis and as illustrated in Table \ref{tab:example_messages1}, GPT-4 and GPT-4-Turbo effectively utilized context in their justifications for their sentiment predictions. This is evidenced by their use of relevant word spans and their correct explanations of the meanings of phrases, leading to accurate predictions and coherent interpretation. GPT-3.5-Turbo lagged slightly due to occasional oversights in contextual (phrase) information. The remaining models including Llama-2-7b, Mistral-7b, Mixtral-8x7b and Gemma-7b often used word-level  rather than phrase level justifications, especially in multilingual and code-mixed texts, leading to misinterpretation of meaning and incorrect predictions or justification.  With regards to cultural relevance, the LLMs generally struggled to incorporate cultural nuances in the dataset, see example in Table \ref{tab:example_messages2}. However, in specific scenarios, models like OpenAI's GPT-4s, which excel in grasping linguistic subtleties and leveraging contextual information, demonstrated proficiency in incorporating cultural aspects into their interpretations.

\textbf{Are the models fluent in maintaining consistency in their interpretation across similar sentiment scenarios within the WhatsApp dataset?} Our findings show that models like GPT-4 and GPT-4-Turbo demonstrate high consistency, reliably applying their analytical capabilities in sentiment predictions across both English and non-English language settings. In contrast, other models performed better on English and Neutral sentiments but lagged behind in non-English cases exhibiting less consistency, often varying in their justifications and output even under similar conditions. This inconsistency can lead to unpredictability in performance in complex multilingual environments, highlighting the need for models to have stable and reliable interpretation mechanisms when deployed in varied real-world applications.

\textbf{Do the models' predictions and interpretation conform to typical human expectation?} We observed that LLMs such as OpenAI's GPT-4s, which demonstrated robust linguistic and contextual comprehension in sentiment detection for the dataset, often aligned with human expectations, consistently producing accurate predictions along with correct justifications. However, the rest of the LLMs, while mostly aligned in English contexts, did not perform as well in non-English settings.

\textbf{Were the models transparent in their decision-making process in the WhatsApp dataset?} Our analysis revealed that Mistral and Mixtral models, despite achieving high F1 scores, often lacked correct and transparent interpretations in their justifications. This was particularly evident in non-English contexts, reflecting issues we observed with LLMs such as GPT-3.5-Turbo, Llama-2-7b, and Gemma-7b. Frequently, these LLMs incorrectly translated the text or failed to disclose the specific text spans that informed their conclusions. This opacity makes it challenging to trust their judgments.  That they often reached the correct prediction could be because the sentiment could be determined despite the incorrect translation (assuming some correlation between the prediction and the justification). The implications of this are that these models might well be useful for tasks like sentiment prediction, but less useful for tasks which require a more accurate interpretation of  language and semantics - including translation tasks.
In contrast, OpenAI's GPT-4s showcased a robust multilingual and contextual performance. These models consistently highlighted relevant phrases or text spans that justified their conclusions, generally leading to correct and transparent outcomes. However, when cultural factors influenced their decision-making process, the models' performance were mixed, displaying transparent interpretation in some instances but falling short in others. This inconsistency raises concerns about the models' reliability in culturally nuanced scenarios.

\begin{table*}[!h]
    \centering 
    \tiny 
    \begin{tabularx}{\textwidth}{*{8}{X}}
        \toprule
        \textbf{Message} & \textbf{GPT-4} & \textbf{GPT-4-Turbo} & \textbf{GPT-3.5-Turbo} & \textbf{Llama-2-70b} & \textbf{Mistral-7b} & \textbf{Mixtral-8x7b} & \textbf{Gemma-7b}\\
        \midrule
        \textbf{Original:} \textit{``Hi guys meeting yetu imekuewaje.''}
        
        \textbf{Translation:} \textit{``Hey guys how was the meeting.''}
        
        \textbf{Actual Sentiment:} Neutral
        & 
        The message `Hi guys meeting yetu imekuewaje' is asking about the status of a meeting, which is a neutral inquiry and does not express any positive or negative sentiment.
        
        \textbf{Prediction:} Neutral
         & 
        The message `Hi guys meeting yetu imekuewaje' is a neutral sentiment as it is a general inquiry about the status or outcome of a meeting without any explicit positive or negative emotion or opinion.

        \textbf{Prediction:} Neutral
        &
        The sentiment is Neutral. The message is a simple greeting and does not express a positive or negative sentiment.

        \textbf{Prediction:} Neutral
        &
        The word `meeting' suggests a gathering or event, which is typically a positive or neutral activity. The phrase `imekuewaje' is in Swahili and roughly translates to `it's okay' or `it's all good' in English, which implies a positive or neutral sentiment. Therefore, the overall sentiment of the message is likely Neutral.

        \textbf{Prediction:} Neutral
        &
        The message is a Swahili greeting, which is a normal aspect and does not express any particular sentiment.

        \textbf{Prediction:} Neutral
        &
        The given message `Hi guys meeting yetu imekuewaje' is a greeting and an invitation to a meeting. The word `guys' is a friendly way to refer to a group of people, `hi' is a common greeting, and `yetu' means `our' in Swahili, indicating inclusivity. The word `imekuewaje' is in Sheng, a slang language in Kenya, and it means `has been organized' or `has been set'. Therefore, the sentiment expressed in this message is Neutral, as it is a factual statement and a simple greeting without expressing any positive or negative emotions.

        \textbf{Prediction:} Neutral
        &
       The phrase `Hi guys meeting yetu imekuewaje' is neutral. The speaker is simply greeting the recipients and does not express any positive or negative sentiment.

        \textbf{Prediction:} Neutral
        \\
        \bottomrule
    \end{tabularx}
    \caption{In this example, only GPT-4 and GPT-4-Turbo showed excellence in both linguistic and contextual understanding while remaining models misinterpreted the message as simply a greeting, due to a misunderstanding of the Swahili words. Interestingly all LLMs made the correct prediction, and their explanations correlated with the prediction they made, but only the explanations by GPT-4 and GPT-4 Turbo matches human expectations.}
    \label{tab:example_messages1}
\end{table*}

\begin{table*}[!h]
    \centering 
    \tiny 
    \begin{tabularx}{\textwidth}{*{8}{X}}
        \toprule
        \textbf{Message} & \textbf{GPT-4} & \textbf{GPT-4-Turbo} & \textbf{GPT-3.5-Turbo} & \textbf{Llama-2-70b} & \textbf{Mistral-7b} & \textbf{Mixtral-8x7b} & \textbf{Gemma-7b}\\
        \midrule
        \textbf{Original:} \textit{``Kama hauko shule shindaapo''}
        
        \textbf{Translation:} \textit{``If you are not in school stay there''}
        
        \textbf{Actual Sentiment:} Negative
        & 
        The message `Kama hauko shule shindaapo' translates to `If you are not in school, stay there'. This is a neutral statement as it does not express any positive or negative sentiment. It is simply a factual advice or suggestion.
        
        \textbf{Prediction:} Neutral
         & 
        The message `Kama hauko shule shindaapo' translates to `If you are not at school, stay where you are' in English. This message seems to be giving neutral advice or instruction without any explicit positive or negative emotion.

        \textbf{Prediction:} Neutral
        &
        The sentiment is Negative. The phrase `shindaapo' expresses a negative feeling or view about not being in school.

        \textbf{Prediction:} Negative
        &
        The sentence `Kama hauko shule shindaapo' expresses a negative sentiment. The word `kama' means `like' in Swahili, and `hauko' means `there isn't' or `there is no'. The phrase `shule shindaapo' means `school is closed'. Therefore, the speaker is expressing a negative sentiment towards the fact that school is closed. The tone of the sentence is also melancholic, indicating a negative emotion.

        \textbf{Prediction:} Negative
        &
        The given query in Swahili translates to `If school had not closed'. It is a statement expressing a factual situation and does not contain any positive or negative sentiment.

        \textbf{Prediction:} Neutral
        &
        The query `Kama hauko shule shindaapo' is in Swahili and translates to `If you hadn't gone to school, you would have passed'. The sentence does not express any positive or negative sentiment. It is a hypothetical situation and there is no sentiment involved. Therefore, the sentiment is Neutral.

        \textbf{Prediction:} Neutral
        &
       The phrase `Kama hauko shule shindaapo' is a Swahili phrase that translates to `You are not going to school today'. The speaker is expressing a negative sentiment as they are saying that the recipient will not be able to attend school. The words `shule' and `shindaapo' are the key words used to convey this sentiment.

        \textbf{Prediction:} Negative
        \\
        \bottomrule
    \end{tabularx}
    \caption{The Swahili phrase carries a negative sentiment as it is used to dismissively suggest that someone who is not in school should refrain from participating in discussions, reflecting a cultural emphasis on educational status as a basis for contributing to intellectual conversations. All LLMs did not understand the cultural context here.}
    \label{tab:example_messages2}
\end{table*}
\section{Related Works} 

\paragraph{Multilingualism in LLMs:} LLMs' effectiveness in multilingual contexts continually attracts attention. Studies by \cite{lai-etal-2023-chatgpt, ahuja-etal-2023-mega, Hendy2023HowGA, Ahuja2023MEGAVERSEBL}, have thoroughly evaluated LLMs across various languages using standard NLP benchmarks with \cite{ojo2023good, robinson-etal-2023-chatgpt} focusing on African languages. However, these studies typically depend on NLP benchmarks that may not closely reflect the complexity of real-world contexts, often missing subtleties and scenarios that occur in everyday use. Moreover, the LLMs might have already encountered some benchmark content online, a concern highlighted by \cite{sainz-etal-2023-nlp, Ahuja2023MEGAVERSEBL}.

\paragraph{Models' Challenges in Code-Mixed Text Processing:} Code-mixing - the blending of two or more languages within a single utterance of a conversation \cite{inbook}, is common in multilingual communities, including Kenya \cite{Orao2012TheKI}. The phenomenon of code-mixing presents unique challenges in the field of NLP. \cite{zhang-etal-2023-multilingual, Doruz2023ASO, Kaji2023ContextualCS} emphasizes the lack of training data as one of the main challenges, attributing to the complexity of processing code-mixed language.

\paragraph{Importance of Real-World Data in LLMs Evaluation:} \cite{Wibowo2023COPALIDIL} introduces COPAL-ID, a culturally rich Indonesian dataset that challenges even advanced models like GPT-4, highlighting the need for nuanced datasets in LLMs evaluation. 
\cite{Chiu2024CulturalTeamingAI} present CulturalTeaming, an AI-assisted interactive red-teaming approach that enhances the creation of multicultural evaluation datasets, revealing significant gaps in LLMs' understanding of diverse cultural contexts through the development of the challenging CULTURALBENCH-V0.1 dataset. 
\cite{Zheng2023LMSYSChat1MAL} curates LMSYS-Chat-1M, a dataset of one million real-world conversations with 25 LLMs, designed to enhance understanding and development of LLMs capabilities in diverse interaction scenarios. Our work extends the efforts on LLMs evaluation using real-world datasets by employing a code-mixed WhatsApp dataset, reflecting a linguistic phenomena absent in curated datasets. Our evaluation combines quantitative and qualitative analysis of LLMs' performance and decision-making processes.
\section{Conclusion}

Our study utilized a multilingual and code-mixed WhatsApp dataset to assess the effectiveness of seven LLMs on a sentiment analysis task. Our evaluation includes both quantitative analysis using metrics like F1 score and qualitative assessment of LLMs' explanations for their predictions. Our comparative analysis revealed that, while Mistral-7b and Mixtral-8x7b achieved high F1 scores, they and other LLMs such as GPT-3.5-Turbo, Llama-2-70b, and Gemma-7b struggled with understanding linguistic and contextual nuances, as well as lack of transparency in their decision-making process as observed from their explanations. In contrast, GPT-4 and GPT-4-Turbo excelled in grasping diverse linguistic inputs and managing various contextual information, demonstrating high consistency with human alignment and transparency in their decision-making process. The LLMs however, encountered difficulties in incorporating cultural nuance especially in non-English settings with the GPT-4s doing so inconsistently. Our evaluation, which leverages real-world data, substantiates the robustness observed in NLP benchmarks, particularly highlighting the superior performance of larger models like OpenAI's GPT-4s in handling low-resource and code-mixed languages. The study highlights the importance of using real-world data for LLMs evaluation. In addition, it advocates for combining qualitative methods from Human-Computer Interaction (HCI) with NLP to gain deeper insights into model performance.

Future research should explore the integration of linguistic diversity and cultural intelligence into model training and evaluation frameworks. Additionally, further work is needed to bridge the gap between quantitative performance metrics and qualitative understandings of model behavior. Lastly, future research should also focus on investigating the relationship between explanations and AI decision-making, for example by quantifying correlations across different NLP tasks. This will ensure that future models are not only effective but also interpretable and aligned with human expectations. 
\section*{Limitations}

This study, though comprehensive, has several limitations. It primarily examines texts in Swahili, English, Sheng, and their code-mixed variants, overlooking the vast array of global languages and dialects. Additionally, the focus on seven specific LLMs provides insights but excludes other emerging LLMs. Lastly, while combining quantitative metrics and qualitative analysis, the balance may constrain the depth of qualitative insights due to the dataset's scale and the subjective nature of qualitative evaluation.

\section*{Ethics Statement}

Given the use of a real-world WhatsApp group chat data, ethical considerations were paramount. All data was anonymized to protect the privacy of individuals, and care was taken to ensure that the dataset did not include sensitive information. The study adhered to ethical guidelines for research involving human subjects, ensuring that all data was used responsibly and respectfully.

% Scientific work published at EMNLP 2022 must comply with the \href{https://www.aclweb.org/portal/content/acl-code-ethics}{ACL Ethics Policy}. We encourage all authors to include an explicit ethics statement on the broader impact of the work, or other ethical considerations after the conclusion but before the references. The ethics statement will not count toward the page limit (8 pages for long, 4 pages for short papers).

% \section*{Acknowledgements}
% The authors express their gratitude to Muchai Mercy, Kemunto Ochwang'i and Faith Ngetich for their invaluable assistance with qualitative analysis of the LLMs' interpretative capabilities.


\clearpage
\bibliography{custom}
\bibliographystyle{acl_natbib}

\appendix
\clearpage
\onecolumn
\section{Appendix}
\label{sec:appendix}

\subsection{The Prompt}
\begin{figure}[h]
\centering
\begin{promptbox}
\justify

You are a helpful NLP assistant, specializing in Sentiment Analysis. You are provided with a WhatsApp chat message (QUERY) in English, Swahili, Sheng, or in more than one language (code-mixed), along with the defintions about the sentiment classes. Your task is to analyze the message and categorize it as Positive, Negative, or Neutral based on the sentiment expressed, along with a justification. Make sure to highlight the words/span of text in the query that you used to make your decision in your justification. \\

\noindent $\langle$DEFINITIONS$\rangle$ \\
\noindent **Negative Sentiment**: It expresses some sort of negative feeling or view or opinion about someone or something. \\

\noindent **Neutral Sentiment**: It neither expresses a positive nor a negative sentiment of the speaker. It could be a general comment, acknowledgement, chitchat or any factual advice or a simple greeting. \\

\noindent **Positive Sentiment**: The sentiment needs to be classified as positive if the speaker feels strong and positive at any particular utterance, except the normal aspects such as any form of greetings. \\
$\langle$/DEFINITIONS$\rangle$ \\ \\

\noindent QUERY: "\{query\}" \\ \\

\noindent \{output\_format\_instructions\} \\
\noindent **DO NOT OUTPUT ANYTHING OTHER THAN THE JSON OBJECT**
\end{promptbox}
\caption{LangChain prompt for Sentiment Analysis. We randomize the order of the definitions to alleviate position bias.}
\label{fig:prompt_here}
\end{figure}

\subsection{Description of Human Evaluation Criteria}

In Table \ref{tab:CriteriaDescription}, we provide a brief description of each of the five rubrics for human evaluation we adopted as outlined by \cite{Chang2023ASO} on \textit{`how to evaluate'}.

 \begin{table*}[h]
     \centering
     \small
     \begin{tabular}{|m{5cm}|m{10cm}|}
     \toprule
    \textbf{Evaluation Criteria} &  \textbf{Description} \\
    \midrule
    Linguistic accuracy & LLM's capacity for precise linguistic interpretation and generation, covering grammar, vocabulary, idioms, and language-specific nuances, while ensuring factual accuracy. \\
    \midrule
    Contextual and cultural relevance & LLM's ability to provide contextually and culturally relevant justifications in sentiment analysis, ensuring responses are appropriate and significant to the given context.  \\
    \midrule
    Fluency in maintaining consistency & LLM's fluency in producing consistent and logical justifications across various sentiment analysis cases, ensuring smooth content flow and uniform tone. \\
    \midrule
    Alignment with human expectations & LLM's ability to produce justifications aligned with human reasoning ensures ethically appropriate predictions, reflecting human values and societal norms, fostering trust in sensitive applications like sentiment analysis. \\
    \midrule
    Transparency in LLM's decision-making process	& LLM's ability to clearly and openly communicate its decision-making process, enabling users to understand the rationale behind responses and gain insights into its inner workings. \\

  \bottomrule
     \end{tabular}
     \caption{Description of Human Evaluation Criteria.}
     \label{tab:CriteriaDescription}
 \end{table*}

\onecolumn
\subsection{LLMs' F1 Scores Across Languages}

\begin{figure*}[ht!]
    \centering
    \includegraphics[width=\textwidth]{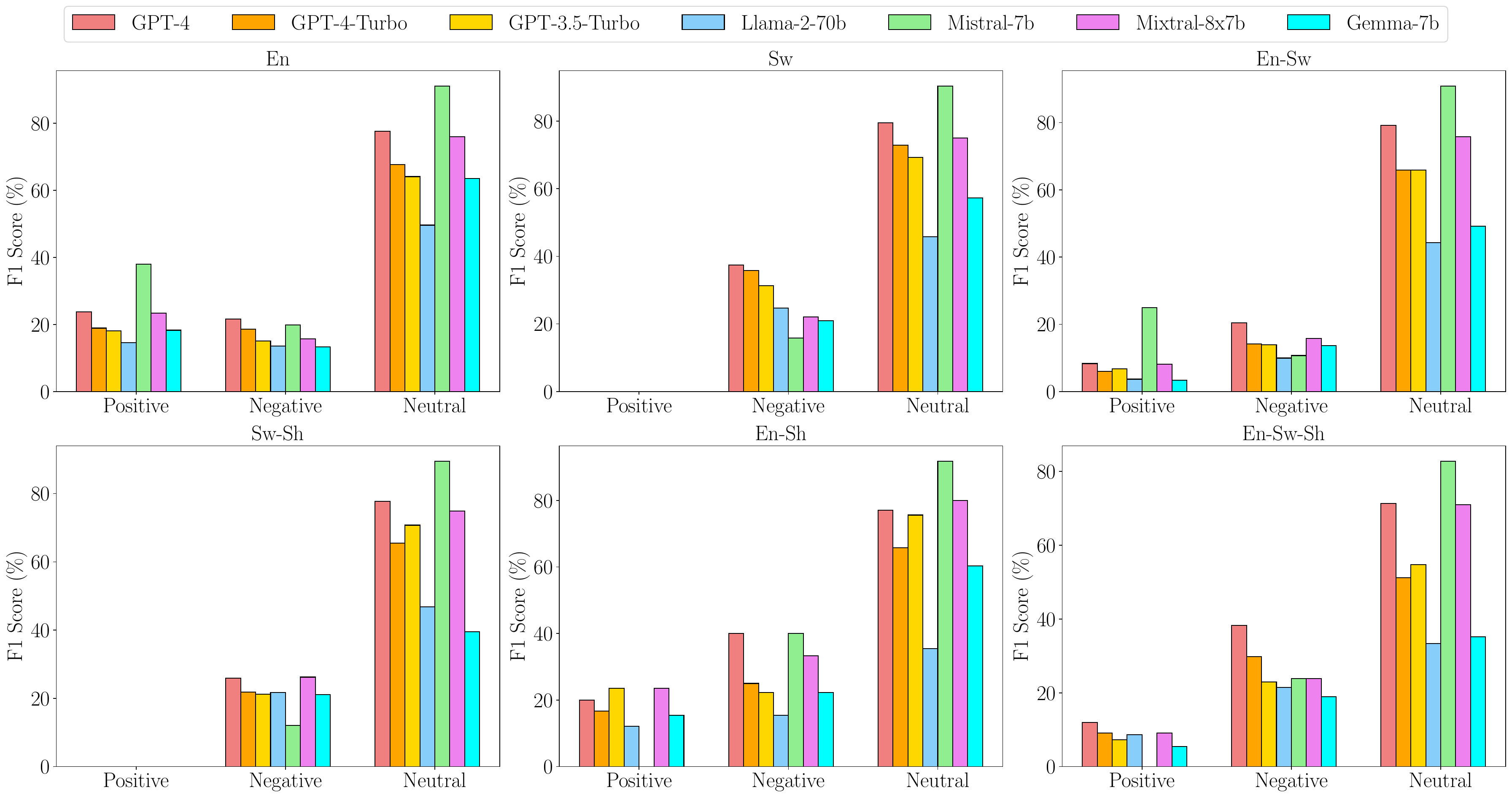}
    \caption{Models' F1 scores across languages for Positive, Negative, and Neutral sentiments.}
    \label{fig:f1_scores_by_languages}
\end{figure*}

\onecolumn
\newpage
\subsection{Detailed Qualitative Analysis of Model Explanations Across Multilingual Examples}
\label{sec:appendix_analysis}

\begin{table*}[!h]
    \centering 
    \tiny 
    \begin{tabularx}{\textwidth}{*{8}{X}}
        \toprule
        \textbf{Message} & \textbf{GPT-4} & \textbf{GPT-4-Turbo} & \textbf{GPT-3.5-Turbo} & \textbf{Llama-2-70b} & \textbf{Mistral-7b} & \textbf{Mixtral-8x7b} & \textbf{Gemma-7b}\\
        \midrule
        \textbf{Original:} \textit{``Mm nimeboeka acha ni lale gud 9t''}
        
        \textbf{Translation:} \textit{``Am bored let me sleep ..goodnight''}
        
        \textbf{Actual Sentiment:} Negative
        & 
        The phrase `nimeboeka' translates to `I'm bored' in English, which indicates a negative sentiment. The speaker is expressing dissatisfaction or a lack of interest, which is a negative feeling.
        
        \textbf{Prediction:} Negative
         & 
        The phrase `nimeboeka' which means `I'm bored' indicates a negative feeling or mood. The speaker's intention to go to sleep (`acha ni lale') could be seen as a way to escape this boredom, which further emphasizes the negative sentiment. The addition of `gud 9t' (good night) does not change the overall sentiment as it is a common phrase used before going to sleep and does not express a strong positive feeling as per the definitions provided.

        \textbf{Prediction:} Negative
        &
        The sentiment is categorized as Neutral. The message does not express a strong positive or negative feeling. It seems to be a casual comment about feeling tired and going to sleep. The use of `acha ni lale gud 9t' indicates the speaker's intention to go to sleep, which is a neutral statement.

        \textbf{Prediction:} Neutral
        &
        The message expresses a positive sentiment due to the use of positive words such as `Mm', `ni lale', and `gud'.

        \textbf{Prediction:} Positive
        &
        The given query is in Swahili language and translates to `I have not found it yet good 9t'. The sentiment expressed is neutral as the speaker is simply stating a fact about their current situation.

        \textbf{Prediction:} Neutral
        &
        The query `Mm nimeboeka acha ni lale gud 9t' is in Sheng, a slang language spoken in Kenya. The phrase roughly translates to `I have given up, it's late now, 9 o'clock'. The sentiment expressed in this message is Negative as the speaker expresses giving up and the lateness of the time could indicate a negative feeling or situation. The word `acha' (give up) is particularly indicative of a negative sentiment.

        \textbf{Prediction:} Negative
        &
       The phrase `Mm nimeboeka acha ni lale gud 9t' is a Sheng phrase that translates to `I am not feeling well'. The speaker is expressing negative sentiment due to the use of the word `acha' and `gud', which are both words used to describe feeling unwell. The phrase also includes a reference to the number `9t', which is commonly used in Sheng to express feeling sick. Therefore, I classify the sentiment of the message as Negative.

        \textbf{Prediction:} Negative
        \\
        \bottomrule
    \end{tabularx}
    \caption{This example involves an English-Swahili message with various textual irregularities including grammatical errors (\textit{`ni lale'} instead of \textit{`nilale'}), local abbreviations (\textit{`Mm'} instead of \textit{`Mimi'}), and misspellings (\textit{`gud 9t'} instead of \textit{`good night'}, \textit{`acha'} instead of \textit{`wacha'}). GPT-4-Turbo excelled in this challenging context by demonstrating its robust multilingual capabilities and resilience to noise, leading to both correct prediction and interpretation. GPT-4 similarly managed to accurately identify the key sentiment indicator in the message. However, the remaining models including GPT-3.5-Turbo, Llama-2-70b, Mistral-7b, Mixtral-8x7b, and Gemma-7b showed weaker contextual and linguistic understanding, as well as a lack of robustness to noise.  As a result, GPT-3.5-Turbo and Mistral-7b made incorrect predictions, while Mixtral-8x7b and Gemma-7b, despite correct predictions, failed in their interpretation.}
    \label{tab:example_messages4}
\end{table*}

\begin{table*}[!h]
    \centering 
    \tiny 
    \begin{tabularx}{\textwidth}{*{8}{X}}
        \toprule
        \textbf{Message} & \textbf{GPT-4} & \textbf{GPT-4-Turbo} & \textbf{GPT-3.5-Turbo} & \textbf{Llama-2-70b} & \textbf{Mistral-7b} & \textbf{Mixtral-8x7b} & \textbf{Gemma-7b}\\
        \midrule
        \textbf{Original:} \textit{``tuache iyo story ju ishaanza kuniboo.''}
        
        \textbf{Translation:} \textit{``Let's leave that story it is starting to bore me.''}
        
        \textbf{Actual Sentiment:} Negative
        & 
        The phrase `ishaanza kuniboo' in the message indicates that the speaker is getting bored or annoyed with the story, which is a negative sentiment.
        
        \textbf{Prediction:} Negative
         & 
        The phrase `ishaanza kuniboo' indicates that the speaker is getting bored or is already bored with the story being discussed. The word `kuniboo' is a Sheng term that translates to 'boring me' in English, which conveys a negative sentiment.

        \textbf{Prediction:} Negative
        &
        The sentiment expressed in the message is negative. The phrase `tuache iyo story' indicates a desire to stop or leave a certain situation, and `ishaanza kuniboo' suggests that something negative has started happening to the speaker. Therefore, the overall sentiment is negative.

        \textbf{Prediction:} Negative
        &
        The word `tuache' means `forget' in Swahili, and the phrase `iyo story' means `that story' or `that thing'. The word `ju' is an interrogative pronoun meaning `what'. Therefore, the overall sentiment of the sentence is Neutral because it is asking about something in a factual manner without expressing any positive or negative emotion.

        \textbf{Prediction:} Neutral
        &
        The given query in Swahili language `tuache iyo story ju ishaanza kuniboo' translates to `leave that story alone' in English. This is a command or instruction and does not express any positive or negative sentiment.

        \textbf{Prediction:} Neutral
        &
        The Swahili phrase `tuache iyo story' translates to `let's leave that story' in English. The word `leave' often indicates a negative sentiment as it suggests abandonment or disinterest. The word `that story' is a reference to a previous conversation or event, which based on the sentiment expressed, seems to be something the speaker wants to move away from. Therefore, the sentiment of this message is Negative.

        \textbf{Prediction:} Negative
        &
       The phrase `tuache iyo story ju ishaanza kuniboo' is a Sheng phrase that translates to `you are telling a story that is not true'. The speaker is expressing disbelief and negativity towards the story being told. The words `ishaanza' and `kuniboo' are the key words used to convey this sentiment.

        \textbf{Prediction:} Negative
        \\
        \bottomrule
    \end{tabularx}
    \caption{This message showcases a complex mix of English, Swahili, and Sheng, with the Swahili-Sheng term \textit{`kuniboo'} serving as a key sentiment indicator. GPT-4 and GPT-4-Turbo excelled in this context, demonstrating strong multilingual capabilities by accurately understanding both the linguistic elements and the contextual nuances, leading to correct predictions and interpretation. GPT-3.5-Turbo and Mixtral-8x7b have made a correct prediction, however, failed to recognize the significance of \textit{`kuniboo'}, likely due to a lack of understanding of this specific term. The remaining models, including Llama-2-70b, Mistral-7b, and Gemma-7b, also missed the critical non-English sentiment indicator. Consequently, Llama-2-70b and Mistral-7b made incorrect predictions with incorrect interpretation, whereas Gemma-7b managed a correct prediction but with flawed interpretation.}
    \label{tab:example_messages3}
\end{table*}

\begin{table*}[!h]
    \centering 
    \tiny 
    \begin{tabularx}{\textwidth}{*{8}{X}}
        \toprule
        \textbf{Message} & \textbf{GPT-4} & \textbf{GPT-4-Turbo} & \textbf{GPT-3.5-Turbo} & \textbf{Llama-2-70b} & \textbf{Mistral-7b} & \textbf{Mixtral-8x7b} & \textbf{Gemma-7b}\\
        \midrule
        \textbf{Original:} \textit{``Send to everyone you love whether it's real love or friendship love, brotherly love, sisterly love, child love, fatherly love and motherly love. You are lovable if you get FIVE sent  back to you !''}
        
        \textbf{Actual Sentiment:} Neutral
        & 
        The message is encouraging the recipient to spread love to everyone they care about, and it also assures them that they are lovable. The phrases `Send to everyone you love' and `You are lovable' indicate a positive sentiment.
        
        \textbf{Prediction:} Positive
         & 
        The message encourages people to send love in various forms and implies that receiving love back is a positive outcome by stating `You are lovable if you get FIVE sent back to you'. The words `love', `lovable', and the concept of sending love to others and receiving it back suggest a positive sentiment.

        \textbf{Prediction:} Positive
        &
        The sentiment expressed in the message is Positive. The speaker is expressing love and positivity towards everyone they care about, including family and friends. The phrase `Send to everyone you love' and the mention of different types of love such as brotherly, sisterly, child, fatherly, and motherly love all contribute to the positive sentiment expressed in the message.

        \textbf{Prediction:} Positive
        &
        The message expresses a positive sentiment as it encourages the reader to spread love and mentions receiving love in return. The use of the word `love' multiple times emphasizes the positive emotion. Additionally, the phrase `you are lovable' suggests that the speaker feels positively about the recipient.

        \textbf{Prediction:} Positive
        &
        The message expresses a positive sentiment towards love and the importance of receiving love in return. The speaker encourages sending love to various relationships and considers being loved as a positive attribute.

        \textbf{Prediction:} Positive
        &
        The message expresses a positive sentiment by encouraging the reader to share love with others, whether it be romantic, familial, or platonic. The phrase `You are lovable if you get FIVE sent back to you' also adds to the positive sentiment as it implies that the recipient is loved and appreciated by others. The use of positive words such as `love', `lovable', and `five' contribute to the overall positive tone of the message.

        \textbf{Prediction:} Positive
        &
       The message expresses strong positive sentiment due to the use of words like `love', `lovable', and `FIVE sent back to you'. The message also emphasizes the feeling of inclusivity and affection towards various relationships.

        \textbf{Prediction:} Positive
        \\
        \bottomrule
    \end{tabularx}
    \caption{This example focuses on the challenges posed by informational-neutral messages that, despite being predominantly in English, contain ambiguous sentiments that may be interpreted as either positive or negative. This ambiguity leads to discrepancies between the models' predictions and the actual annotations. Despite these challenges, the example also highlights all models' strong linguistic understanding of English, showcasing their ability to handle the language even in nuanced scenarios.}
    \label{tab:example_messages5}
\end{table*}

\begin{table*}[!h]
    \centering 
    \tiny 
    \begin{tabularx}{\textwidth}{*{8}{X}}
        \toprule
        \textbf{Message} & \textbf{GPT-4} & \textbf{GPT-4-Turbo} & \textbf{GPT-3.5-Turbo} & \textbf{Llama-2-70b} & \textbf{Mistral-7b} & \textbf{Mixtral-8x7b} & \textbf{Gemma-7b}\\
        \midrule
        \textbf{Original:} \textit{``Me siyo beste yako''}
        
        \textbf{Translation:} \textit{``Am not your friend''}
        
        \textbf{Actual Sentiment:} Negative
        & 
        The phrase `Me siyo beste yako' translates to `I am not your friend' in English. This statement expresses a negative sentiment as it indicates a rejection or negative view towards the person being addressed.
        
        \textbf{Prediction:} Negative
         & 
        The phrase `Me siyo beste yako' in Sheng translates to `I am not your friend' in English. The use of `siyo' (not) indicates a denial or negation of friendship, which conveys a negative sentiment towards the relationship with the person being addressed.

        \textbf{Prediction:} Negative
        &
        The sentiment is Negative. The phrase `siyo beste yako' in Swahili translates to `not your best', which expresses a negative opinion about someone or something.

        \textbf{Prediction:} Negative
        &
        The phrase `Me siyo beste yako' is a Swahili phrase that translates to `I don't have your best interests at heart' in English. The use of the phrase implies that the speaker does not have the listener's well-being in mind, and may even have negative intentions. Therefore, the sentiment of this phrase is Negative.

        \textbf{Prediction:} Negative
        &
        The query `Me siyo beste yako' in Swahili translates to `I am not your best' in English. This statement is a simple negation and does not express any strong positive or negative sentiment. It could be a response to a request for compliments or an expression of humility.

        \textbf{Prediction:} Neutral
        &
        The query `Me siyo beste yako' is in Sheng, a code-mixed language in Kenya. The phrase can be translated to `I am not your best' in English. The use of `not' indicates a negative sentiment. However, the sentiment is not strong as it's a statement of fact rather than an expression of a negative emotion. Therefore, the sentiment is Negative.

        \textbf{Prediction:} Negative
        &
       The phrase `Me siyo beste yako' is a Swahili phrase that translates to `I am your best friend' in English. The phrase expresses a strong positive sentiment. The words `beste' and `yako' are the key words that I used to make this decision.

        \textbf{Prediction:} Positive
        \\
        \bottomrule
    \end{tabularx}
    \caption{This example demonstrates the difficulties all models face when processing the Sheng language, particularly for those with weaker linguistic understanding of non-English languages. Only GPT-4 and GPT-4-Turbo successfully identified and correctly interpreted the language, highlighting the significant variability in multilingual capabilities across the models.}
    \label{tab:example_messages6}
\end{table*}

\end{document}